%% file: icomp2024_conference.tex
\documentclass{article} 
\usepackage{icomp2024_conference,times}

\input{math_commands.tex}

\usepackage{hyperref}
\usepackage{url}
\usepackage{graphicx}
\usepackage{caption}
\usepackage{float}
\usepackage{subcaption}
\usepackage{booktabs} 

\title{Training-Free Out-of-Distribution Segmentation with Foundation Models}

\author{
Laith Nayal$^{1,4}$, 
Hadi Salloum$^{2,3,4}$, 
Ahmad Taha$^{5}$, 
Yaroslav Kholodov$^{2,4}$, 
Alexander Gasnikov$^{2,3}$ \\
\\
$^{1}$ Laboratory of Multimodal Research In Industry, AI Institute, Innopolis University\\
$^{2}$ Phystech School of Applied Mathematics and Computer Science, Moscow Institute of Physics and Technology, \\
Institutsky lane 9, Dolgoprudny, Moscow region, 141700\\
$^{3}$ Research Center for Artificial Intelligence, Innopolis University\\
$^{4}$ Q Deep, Innopolis, 420500, Russia \\
$^{5}$ Machine Learning and Data Representation Lab, Innopolis University, Innopolis, 420500, Russia \\
}

\icompfinalcopy 
\begin{document}

\maketitle

\begin{abstract}
 Detecting unknown objects in semantic segmentation is crucial for safety-critical applications such as autonomous driving. Large vision foundation models, including DINOv2, InternImage, and CLIP, have advanced visual representation learning by providing rich features that generalize well across diverse tasks. While their strength in closed-set semantic tasks is established, their capability to detect out-of-distribution (OoD) regions in semantic segmentation remains underexplored. In this work, we investigate whether foundation models fine-tuned on segmentation datasets can inherently distinguish in-distribution (ID) from OoD regions without any outlier supervision. We propose a simple, training-free approach that utilizes features from the InternImage backbone and applies K-Means clustering alongside confidence thresholding on raw decoder logits to identify OoD clusters. Our method achieves 50.02 Average Precision on the RoadAnomaly benchmark and 48.77 on the benchmark of ADE-OoD with InternImage-L, surpassing several supervised and unsupervised baselines. These results suggest a promising direction for generic OoD segmentation methods that require minimal assumptions or additional data.
\end{abstract}

\section{Introduction}
\label{sec:intro}

The emergence of large vision foundation models, such as DINOv2~\citet{oquab2024dinov2learningrobustvisual}, InternImage~\citet{Wang_2023_CVPR}, and CLIP~\citet{CLIP}, trained on Internet-scale data, has significantly advanced the field of visual representation learning.
These models provide strong features that can generalize across diverse tasks, substantially reducing the reliance on task-specific supervision. More specifically, their features have demonstrated significant improvements in downstream tasks that assume a closed set of semantic concepts, such as detection and segmentation~\citet{Wang_2023_CVPR}. Therefore, in this work, we aim to explore whether the strong generalization ability of foundation models can allow expressing uncertainty and detecting OoD samples for the downstream tasks they are used for.



Existing OoD segmentation methods can be broadly categorized into two classes based on their use of training data. The first class includes methods that rely on \emph{outlier exposure}, where additional OoD data is used during training. Examples include RbA~\cite{Nayal_2023_ICCV} and PEBAL~\cite{PEBAL}, which train models to assign lower confidence to outlier regions by modeling them as a distinct class.

The second class includes methods that operate without any additional outlier data. These approaches rely solely on the model's behavior learned from in-distribution data~\cite{sml}. In practice, segmentation models often exhibit increased uncertainty over OoD regions, even without explicit supervision. However, their raw logit outputs are typically poorly calibrated and may lack consistency in their responses to unseen inputs.

\vspace{0.5em}
\noindent\textbf{Objective.}  
The goal of this work is to explore the ability of large vision foundation models which are fine-tuned on semantic segmentation datasets, such as Cityscapes~\cite{Cordts_2016_CVPR} and ADE-OoD~\cite{DooD}, to distinguish ID from OoD regions without relying on any outlier supervision. We argue that the robustness and richness of the foundation model features provide a sufficient signal to semantically distinguish ID from OoD regions, without any signals from pseudo-OoD finetuning.

\begin{figure}[t]
    \centering
    \includegraphics[width=0.43\textwidth]{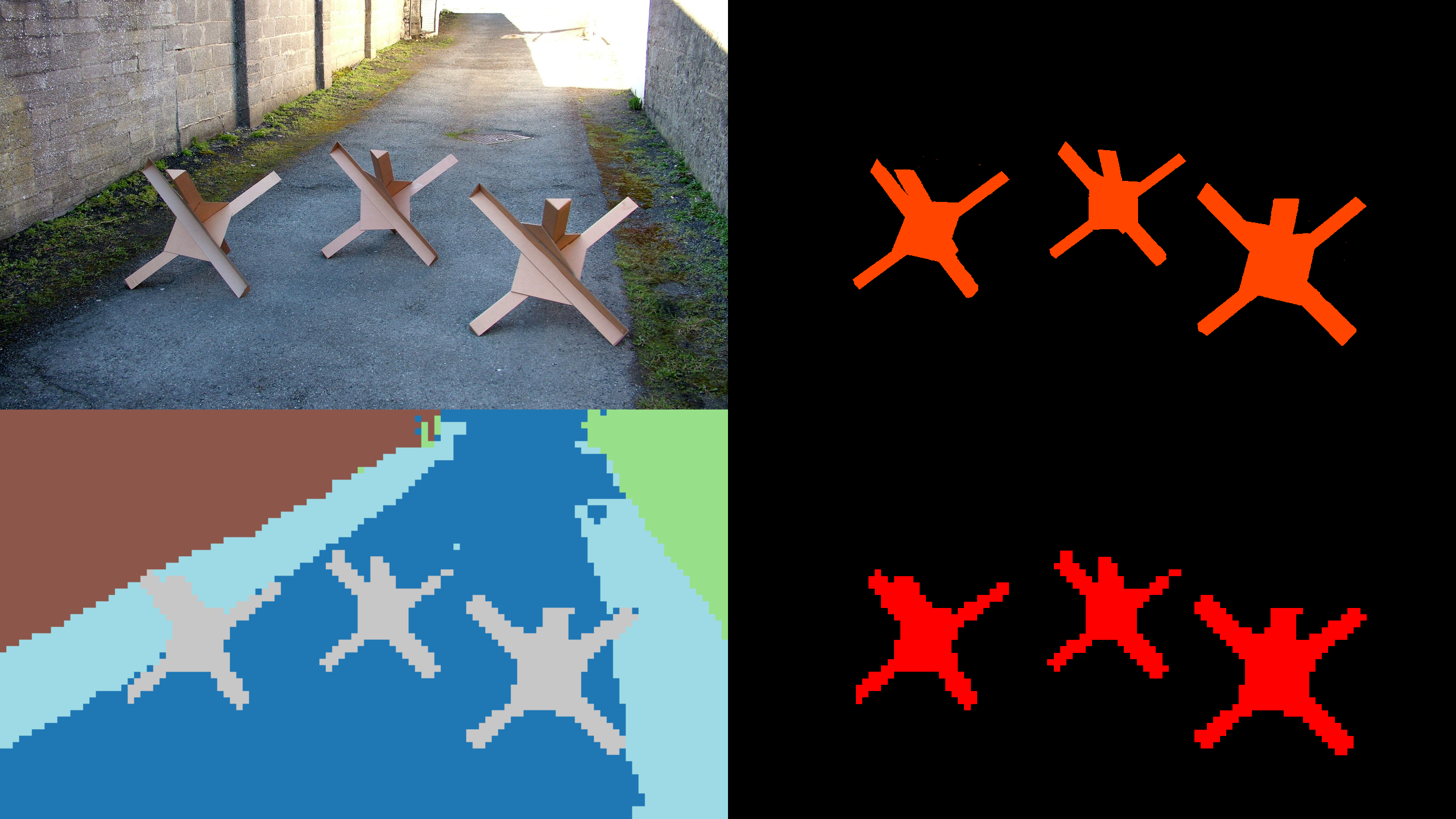}
    \caption{Input image (top-left), Ground Truth (top-right), Clusters (bottom-left), Ours (bottom-right).}
    \label{fig:ood_results}
\end{figure}

\section{Related Work}

\textbf{Foundation Models for OoD:} Vision foundation models have demonstrated a great zero-shot performance on downstream tasks such as classification and segmentation. This high performance has resulted in a noticeable boost in Out-of-Distribution (OoD) detection performance, moving away from classical approaches---like \textit{ensembles}~\cite{NIPS2017_9ef2ed4b} and \textit{Monte Carlo dropout}~\cite{pmlr-v48-gal16}, which require multiple forward passes and are therefore computationally expensive in practice. Recently, state-of-the-art approaches leverage foundation models as frozen backbones for feature extraction, taking advantage of their semantically rich representations.

Recent works have utilized DINOv2~\cite{oquab2024dinov2learningrobustvisual}, a self-distilled vision transformer, in methods such as \textit{PixOOD}~\cite{PixOOD} and \textit{UEM}~\cite{nayal2024likelihoodratiobasedapproachsegmenting}, and others have adopted CLIP~\cite{CLIP} as a backbone for OoD detection. In contrast, our study explores the \textit{InternImage} model~\cite{Wang_2023_CVPR}, which, to the best of our knowledge, has not yet been investigated in the context of OoD detection. InternImage is the first large-scale CNN-based model to incorporate flexible convolution variants---specifically, \textit{deformable convolutions}~\cite{Dai_2017_ICCV,Zhu_2019_CVPR}---which enable dynamic sampling offsets and improve the learning of receptive fields. The model achieves state-of-the-art performance on semantic segmentation benchmarks, particularly on datasets such as Cityscapes~\cite{Cordts_2016_CVPR} (see Supplementary Materials).

\textbf{No Additional Training:} Some approaches do not involve any additional training. These methods often use the model’s prediction confidence from semantic segmentation models as a proxy for OoD scores. For example, some methods utilize the predicted softmax distributions to measure uncertainty, such as \textit{Maximum Softmax Probability (MSP)}~\cite{msp}. However, in practice, utilizing the raw logits before softmax yields better results. In this context, the paper \textit{Standardized Max Logits (SML)}~\cite{sml} proposes standardizing the maximum logits in a class-wise manner to align distributions. While this approach introduces negligible computational overhead, its performance is limited on several anomaly detection benchmarks.

Another method, \textit{Mask2Anomaly}~\cite{Mask2Anomal}, belongs to the category of approaches that freeze the segmentation model and classify OoD pixels based on distance measures. It computes a confidence score using the \textit{Mahalanobis distance} and does not require additional training or external OoD data.

A notable technique is the \textit{Dense Nearest-Neighbor-Based Out-of-Distribution Detection (cDNP)}~\cite{galesso2023farawaydeepspace}, which leverages features extracted from deep supervised semantic segmentation models and applies \textit{$k$-nearest neighbors (kNN)} to estimate density for OoD detection.

\textbf{Outlier Exposure / Supervision:} \textit{Outlier Exposure (OE)} is a popular recent method that involves using an auxiliary dataset of outliers during training. The model is trained to maximize uncertainty for these proxy anomalies, aiming to generalize to unseen outliers. Some state-of-the-art methods such as \textit{Rba}~\cite{Nayal_2023_ICCV} and \textit{PEBAL}~\cite{PEBAL} explicitly train the model to produce lower logit scores for proxy OoD samples.

However, OE-based approaches tend to incur higher computational cost, and in some cases, OE can degrade inlier segmentation performance by biasing the model toward anomaly classes.

\textbf{Beyond the Above Methods:} \textit{Diffusion for Out-of-Distribution Detection (DOoD)}~\cite{DooD} introduces a novel OoD detection approach based on diffusion models. DOoD trains a diffusion model, specifically a small multi-layer perceptron (MLP), on in-distribution semantic embeddings to reconstruct features while discarding harmful spatial correlations. At inference time, it uses score matching to compute pixel-wise OoD scores. Although it does not use OoD data during training, DOoD performs well under large domain shifts and diverse inlier distributions.

\section{Methodology}

\begin{figure*}
    \centering
    \includegraphics[width=\linewidth]{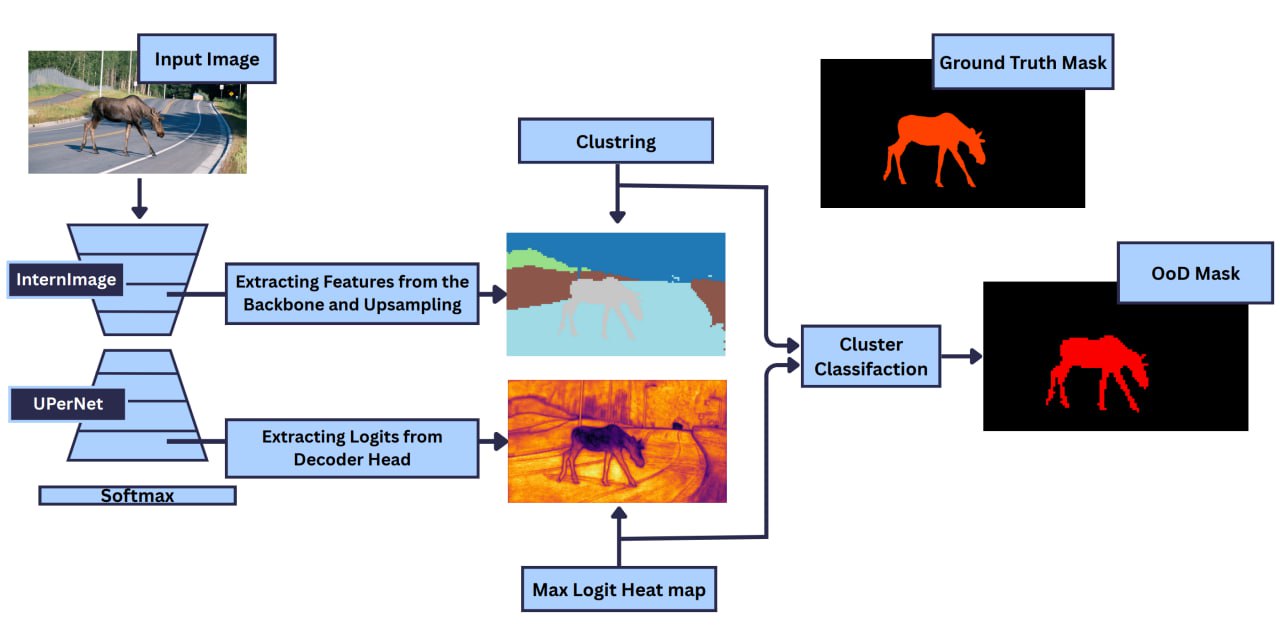}
    \caption{Proposed training-free OoD segmentation pipeline: Input images are processed by the InternImage backbone for feature extraction and upsampling. Simultaneously, UPerNet decodes semantic logits, generating max-logit confidence heatmaps. Features are clustered via K-Means, and cluster classification identifies OoD groups (gray cluster). Final OoD masks are compared against ground truth.}
    \label{fig:enter-label}
\end{figure*}
In this work, we propose a simple and training-free method to predict Out-of-Distribution (OoD) masks.
\subsection{Feature Extraction}

\textbf{Feature Extraction:} To extract features, we forward the input samples through the backbone of the InternImage model~\cite{Wang_2023_CVPR}, which is composed of four hierarchical downsampling blocks with deformable convolution layers~\cite{Dai_2017_ICCV,Zhu_2019_CVPR}. We extract features from the third block, as it produces fine-grained and semantically rich representations, which are especially suitable for clustering-based segmentation (For more information about InternImage, check Supplementary materials).

Let $x \in \mathbb{R}^{H \times W \times 3}$ denote the input image, and $f(x) \in \mathbb{R}^{H' \times W' \times D}$ the feature map from the third block, where $H' < H$, $W' < W$, and $D$ is the feature dimensionality.

\subsection{Clustering}

\paragraph{Clustering:} We apply $K$-Means clustering~\cite{macqueen1967kmeans} on the extracted features $f(x)$. While $K$-Means assumes a fixed number of clusters $K$, which may be limiting, it remains computationally efficient and effective for feature-space partitioning.



\subsection{Logits Extraction}

\textbf{Logit Extraction:} We extract the raw class logits from the decoder head UPerNet~\cite{xiao2018unified} fine-tuned on cityscapes~\cite{Cordts_2016_CVPR}. Let $z \in \mathbb{R}^{H \times W \times C}$ denote the logit tensor, where $C$ is the number of semantic classes, and $(H, W)$ is the spatial resolution of the prediction map. These logits are obtained from the last layer prior to softmax.

\subsection{Max Logits for Confidence Estimation}

\textbf{Extracting Max Logits:} To estimate model confidence at each pixel, we compute the maximum logit value across the class dimension:

\begin{equation}
m(x) = \max_{c \in \{1,\ldots,C\}} z_{x,c},
\end{equation}

where $z_{x,c}$ is the logit value at pixel $x$ for class $c$. Higher values of $m(x)$ typically indicate greater model confidence, while lower values suggest uncertainty and potential OoD regions.

\subsection{Cluster Classification Using Logits}

\textbf{Cluster-Based Classification:} After obtaining the clustered feature map and the logit confidence map $m(x)$, we align their spatial dimensions via bilinear interpolation. Specifically, we upsample the feature clusters to match the resolution of the logits:

\begin{equation}
\tilde{f}(x) = \text{Upsample}(f(x)) \in \mathbb{R}^{H \times W}.
\end{equation}

Next, for each cluster $k$, we compute the proportion of pixels whose confidence falls below a threshold $\tau$:

\begin{equation}
r_k = \frac{1}{|\mathcal{C}_k|} \sum_{x \in \mathcal{C}_k} \mathbb{1}[m(x) < \tau],
\end{equation}

where $\mathcal{C}_k$ is the set of pixels in cluster $k$, and $\mathbb{1}[\cdot]$ is the indicator function.

We declare cluster $k$ as OoD if $r_k > T$, where $T \in [0,1]$ is a predefined ratio threshold.

\section{Experiments and Analysis}

\begin{table}[t]
\centering
\small
\setlength{\tabcolsep}{6pt} 
\begin{tabular}{c l c c c}
\toprule
\textbf{Rank} & \textbf{Model} & \textbf{Training Free} & \textbf{AP ↑} & \textbf{FPR95 ↓} \\
\midrule
1  & OodDINO                           & No & 95.21 &  2.11 \\
2  & RbA~\cite{Nayal_2023_ICCV}        & No & 90.28 &  4.92 \\
3  & DOoD~\cite{DooD}                  & No  & 89.10 &  8.80 \\
4  & cDNP~\cite{galesso2023cDNP}       & Yes  & 85.60 &  9.80 \\
5  & Mask2Anomaly~\cite{Mask2Anomal}   & No & 79.70 & 13.45 \\
6  & RPL+CoroCL~\cite{Liu_2023_ICCV}   & No & 71.61 & 17.74 \\
7  & \textbf{InternImage-L (ours)}     & Yes  & \textbf{50.02} & \textbf{61.55} \\
8  & PEBAL~\cite{PEBAL}                & Yes & 45.10 & 44.58 \\
9  & Synboost~\cite{Di_Biase_2021_CVPR}& Yes & 41.83 & 59.72 \\
10 & \textbf{InternImage-B (ours)}     & Yes  & \textbf{37.30} & \textbf{70.38} \\
11 & SML~\cite{sml}                    & Yes & 25.82 & 49.74 \\
12 & SynthCP~\cite{xia2020synthesize}  & Yes & 24.86 & 64.69 \\
\bottomrule
\end{tabular}
\caption{OoD segmentation results on the Road Anomaly dataset. Models are ranked by Average Precision (AP). We also report the False Positive Rate at 95\% True Positive Rate (FPR95). Models that do not rely on additional data are highlighted, and our results are marked as (ours).}
\label{tab:ood_results}
\end{table}

\begin{table}[t]
\centering
\small
\setlength{\tabcolsep}{6pt} 
\begin{tabular}{c l c c c}
\toprule
\textbf{Rank} & \textbf{Model} & \textbf{Training-free} & \textbf{AP ↑} & \textbf{FPR95 ↓} \\
\midrule
1  & RbA~\cite{Nayal_2023_ICCV}             & No & 66.82 & 82.42 \\
2  & DOoD~\cite{DooD}                        & No  & 63.03 & 36.50 \\
3  & cDNP~\cite{galesso2023cDNP}            & Yes  & 62.35 & 39.20 \\
4  & \textbf{InternImage-L (ours)}           & Yes  & \textbf{48.77} & \textbf{69.36} \\
5  & GMMSeg~\cite{liang2022gmmseg}                    & No  & 47.60 & 43.50 \\
6  & \textbf{InternImage-B (ours)}           & Yes  & \textbf{33.09} & \textbf{72.17} \\
7  & Maskomaly~\cite{ackermann2023maskomaly}              & No  & 28.23 & 82.68 \\
8  & M2A~\cite{Mask2Anomal}                          & No  & 18.23 & 86.53 \\
\bottomrule
\end{tabular}
\caption{ADE-OoD dataset comparison, sorted by Average Precision (AP) in descending order. AP ↑ indicates higher is better, FPR95 ↓ indicates lower is better.}
\label{tab:ade_ood_results_sorted}
\end{table}

\subsection{Dataset}

In our preliminary experiments, we use the \textit{RoadAnomaly} benchmark~\cite{Lis_2019_ICCV}, which is an earlier and smaller version of the \textit{SegmentMeIfYouCan (SMIYC)} benchmark~\cite{chan2021segmentmeifyoucan}. RoadAnomaly consists of 60 images containing diverse objects in various road environments.  
RoadAnomaly assess a model’s ability to segment OoD objects on the road that do not belong to any of the 19 predefined semantic classes in the Cityscapes dataset~\cite{Cordts_2016_CVPR}.

To provide a more comprehensive evaluation, we also report results on the ADE-OoD benchmark. ADE-OoD was constructed using images from the validation set of ADE20k and OpenImages~\cite{kuznetsova2020open}.

\subsection{Experimental Setup}

\textbf{Experimental Setup:} All experiments are performed without additional training or fine-tuning. 

\subsubsection{Model Sizes and Architectures}

\textbf{Models:} We investigate our method using the InternImage~\cite{Wang_2023_CVPR} backbone in two variants:
\begin{itemize}
    \item InternImage-B (Base): 128M parameters
    \item InternImage-L (Large): 256M parameters
\end{itemize}
As the decoder head, we use UPerNet~\cite{xiao2018unified} to obtain high-resolution semantic predictions.

\subsubsection{Hyperparameters}

We conduct experiments to determine effective hyperparameter values for both Cityscapes and ADE-OoD benchmarks.

\textbf{Cityscapes:}  
\begin{itemize}
    \item \textbf{Cluster Count ($K$):} $K = 5$ for $K$-Means clustering. Most images contain 4--5 active semantic categories; an additional cluster accounts for unexpected or OoD content.
    \item \textbf{Confidence Threshold ($\tau$):} $\tau = 1.5$, estimated from the logit distributions in the Cityscapes training set. Pixels with max logits below this value are considered uncertain.
    \item \textbf{Cluster OoD Ratio ($T$):} $T = 30\%$. A cluster is classified as OoD if more than $T$ percent of its pixels fall below the confidence threshold. This value provides the best trade-off between recall and false positives.
\end{itemize}

\textbf{ADE-OoD:}  
\begin{itemize}
    \item \textbf{Cluster Count ($K$):} $K = 6$, estimated from the training set to capture diverse categories and potential OoD regions.
    \item \textbf{Confidence Threshold ($\tau$):} $\tau = 1.1$, estimated from the ADE-OoD training set.
    \item \textbf{Cluster OoD Ratio ($T$):} $T = 40\%$, empirically determined to provide optimal detection performance.
\end{itemize}

\subsection{Evaluation Metrics}

\textbf{Metrics:} For quantitative comparison on the RoadAnomaly benchmark, we report:
\begin{itemize}
    \item \textbf{Average Precision (AP)}: Area under the precision-recall curve for OoD segmentation.
    \item \textbf{False Positive Rate at 95\% True Positive Rate (FPR@95)}: Measures the percentage of ID pixels wrongly classified as OoD when 95\% of OoD pixels are correctly detected.
\end{itemize}

\subsection{Quantitative Results}

\subsubsection{RoadAnomaly}

As shown in table \ref{fig:ood_results}, our results on the RoadAnomaly benchmark, outperforms existing approaches such as \textit{PEBAL}~\cite{PEBAL}, \textit{SynBoost}~\cite{Di_Biase_2021_CVPR}, \textit{Standardized Max Logits (SML)}~\cite{sml}, and \textit{SynthCP}~\cite{xia2020synthesize}. 

Notably, some of these methods rely on additional training or outlier exposure, whereas our approach remains entirely training-free. While our method demonstrates good AP performance, we observe a higher false positive rate. We hypothesize that this may be due to the strong overlap between certain ID and OoD feature distributions in complex scenes, which affects clustering accuracy.

\subsubsection{ADE-OoD}
Table~\ref{tab:ade_ood_results_sorted} reports the performance of our method compared to existing baselines on the ADE-OoD dataset.

Our method demonstrates competitive AP and robust OoD detection compared to prior approaches. Notably, InternImage-L achieves \textbf{48.77 AP} and \textbf{69.36 FPR95}, while InternImage-B achieves \textbf{33.09 AP} and \textbf{72.17 FPR95}, without using any extra data. 

These results are particularly significant because we evaluate our method on two distinct benchmarks: the \textit{RoadAnomaly} dataset~\cite{Lis_2019_ICCV} and the ADE-OoD~\cite{DooD} dataset, which represent two different distribution shifts. Consistent performance across these benchmarks demonstrates that our approach is not overfitted to a single dataset and is robust to variations in scene composition, object categories, and environmental conditions. This validates the effectiveness of our pipeline in detecting OoD objects across diverse real-world scenarios.

Overall, our quantitative analysis confirms that the proposed method is reliable and generalizable, effectively identifying OoD regions even under distribution shifts.

\subsection{Qualitative Results}

We provide diverse qualitative results on the RoadAnomaly benchmark (See Supplementary). The visualizations reveal promising separation between ID and OoD regions.

We experiment with different values of $K$ in K-means clustering (e.g., $K = 4, 5, 6$). When $K$ is small, ID features tend to cluster together and OoD features form a distinct group. With larger $K$, OoD regions may split into multiple clusters. This is especially observable in larger OoD objects, which remain consistently detected across various $K$ values.

\section{Conclusion, Limitations, and Future Work}

From this work, we observe that vision foundation models possess strong visual representations. While our results are not near the state-of-the-art, we believe they are still promising—especially given the simplicity and naivety of our approach (e.g., K-Means clustering and max-logit thresholding), as well as the use of a basic upsampling method. In future work, we plan to adopt more robust upsampling techniques such as FeatUp~\cite{fu2024featup} and JAFAR~\cite{couairon2025jafar}.

Our broader goal is to develop a method or recipe that enables robust OoD segmentation with minimal constraints and broad applicability to existing segmentation architectures. In other words, our aim is to devise a formulation that can convert any closed-set segmentation model into an open-world segmentation model.

The UEM~\cite{nayal2024likelihoodratiobasedapproachsegmenting} method have attempted to address this challenge by proposing modules that can be trained on top of any segmentation model to enable open-world behavior. However, this method still rely on restrictive assumptions and require outlier training data. By carefully analyzing the behavior and failure modes of such approaches, we hope to move toward a more general framework for OoD segmentation.

\subsubsection*{Author Contributions}

\subsubsection*{Acknowledgments}

\bibliography{icomp2024_conference}
\bibliographystyle{icomp2024_conference}

\appendix
\section{Appendix}

\subsection{InternImage}
\label{sec:internimage}

\begin{figure}[H]
    \centering
    \includegraphics[width=0.485\textwidth]{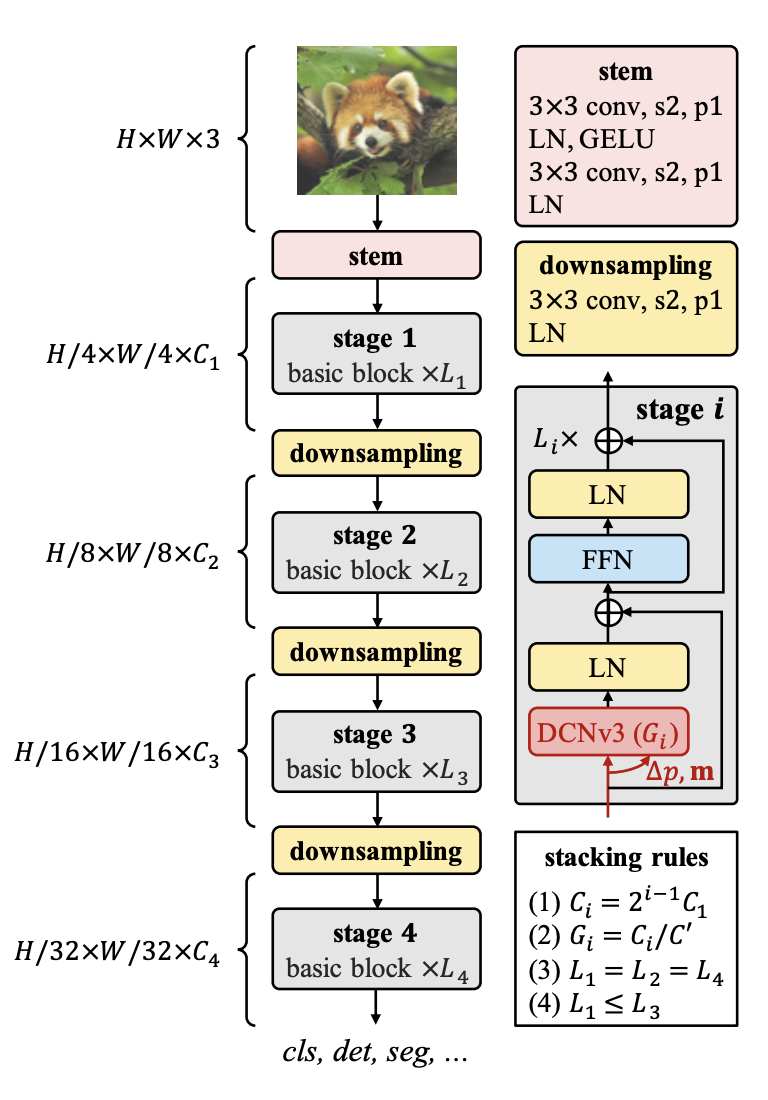}
    \caption{(a) Hierarchical architecture with stem/downsampling layers; 
    (b) DCNv3 block: LN $\rightarrow$ deformable conv $\rightarrow$ FFN; 
    (c) Multi-group sampling adapts to object structure.}
    \label{fig:internimage_arch}
\end{figure}

InternImage bridges the performance gap between \textbf{CNNs and ViTs} via \textbf{Deformable Convolution v3 (DCNv3)}.  
As shown in Fig.~\ref{fig:internimage_arch}, this foundation model delivers ViT-like adaptability with CNN efficiency through three key innovations:

\begin{enumerate}
    \item \textbf{Dynamic sparse sampling} (3$\times$3 kernel)
    \item \textbf{Input-adaptive} spatial aggregation
    \item \textbf{Multi-group} hierarchical features
\end{enumerate}

\begin{table}[h]
\centering\footnotesize
\begin{tabular}{lccc}
\toprule
\textbf{Backbone} & \textbf{mIoU (SS)} & \textbf{mIoU (MS)} & \textbf{Params} \\
\midrule
InternImage-B & 83.18 & 83.97 & 128M \\
InternImage-L & 83.68 & 84.41 & 256M \\
\bottomrule
\end{tabular}
\vspace{-0.2cm}
\caption{Cityscapes results (512$\times$1024) with UperNet. MS: Multi-scale.}
\label{tab:cityscapes}
\end{table}

\section{Supplementary Visual Results}
\label{sec:visual_results}

This section presents qualitative segmentation results on RoadAnomaly dataset.  

\begin{figure}[H]
    \centering
    \includegraphics[width=\textwidth]{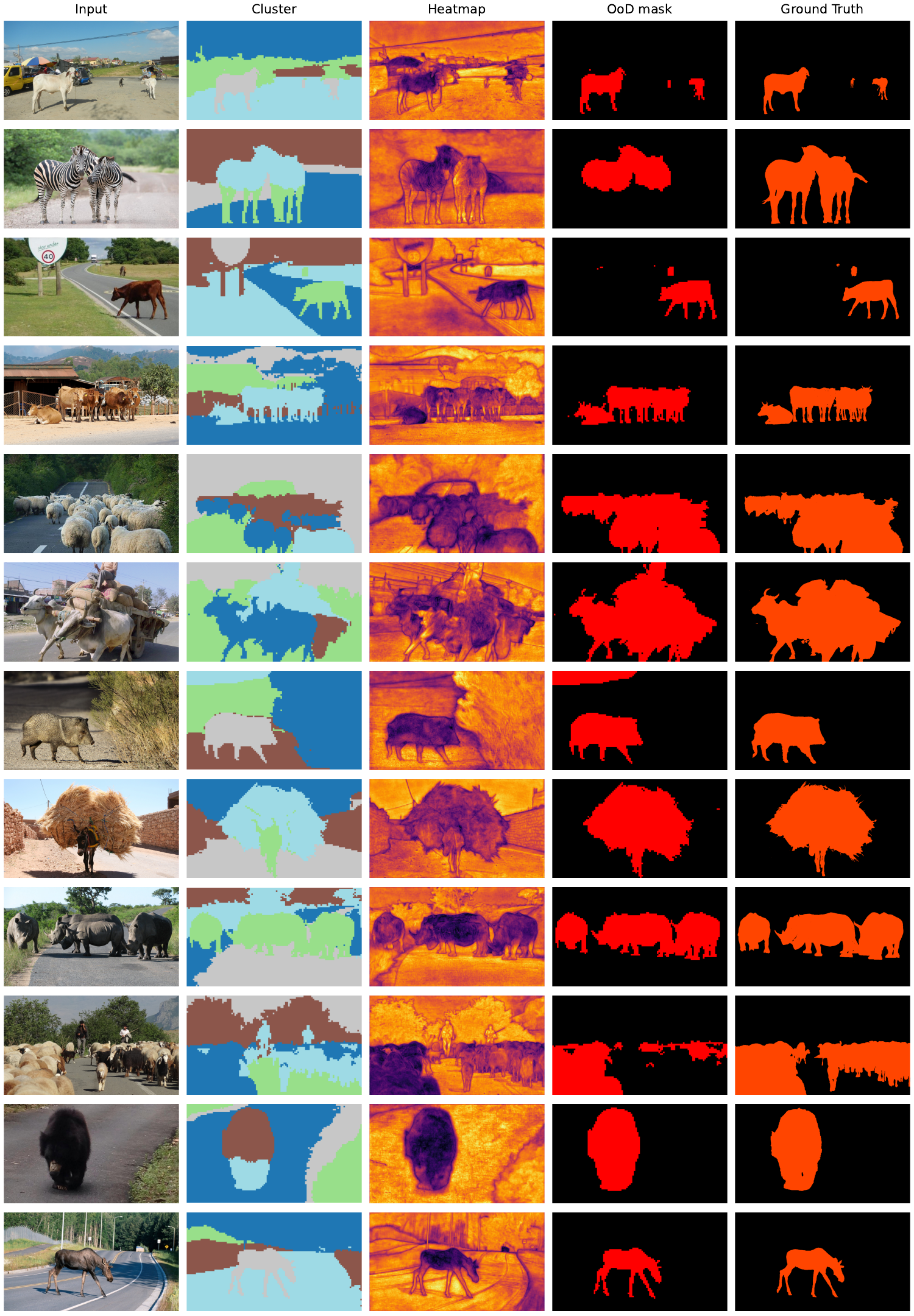}
    \caption{Qualitative segmentation results (Set 1).}
    \label{fig:supp1}
\end{figure}

\clearpage
\begin{figure}[H]
    \centering
    \includegraphics[width=\textwidth]{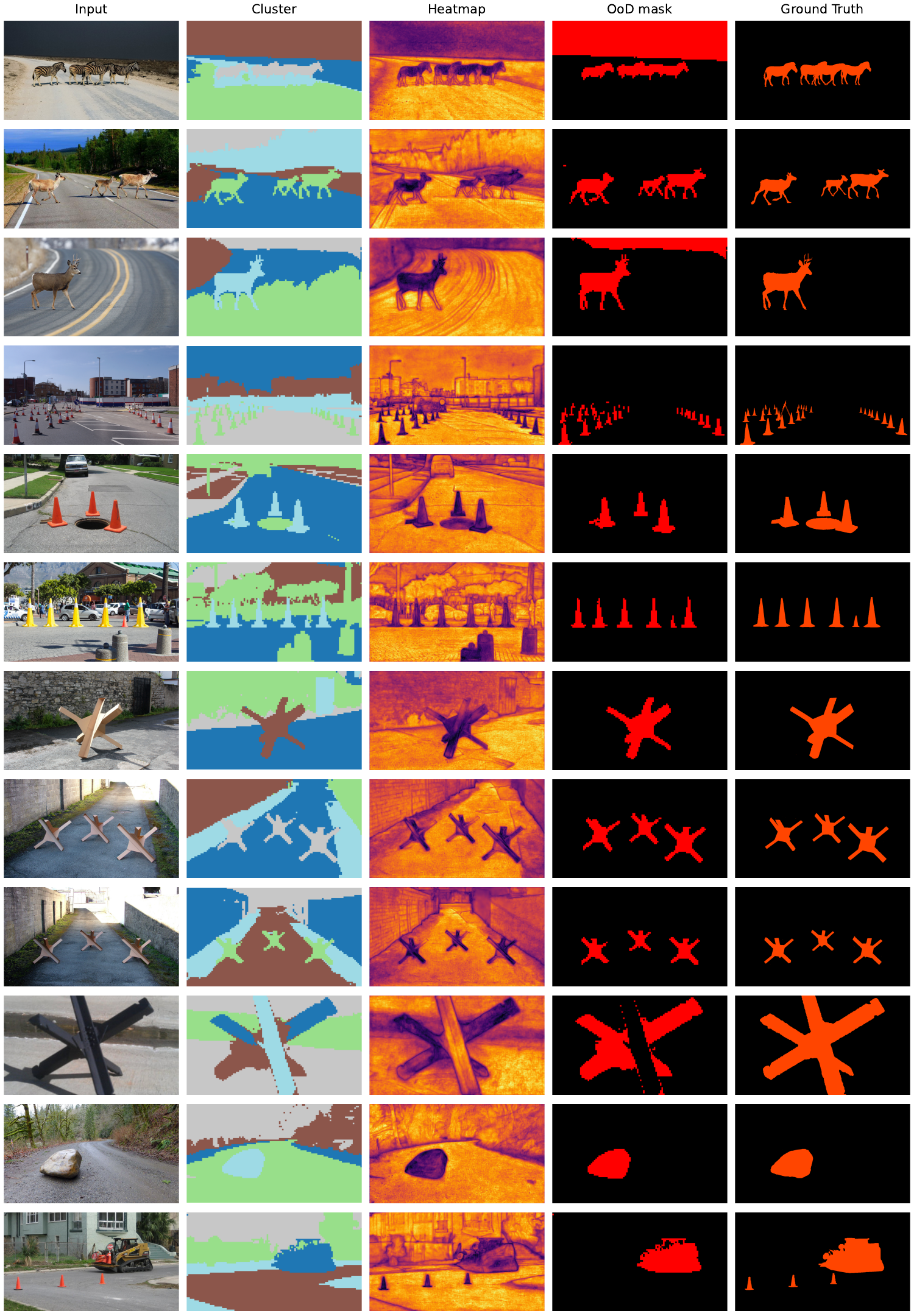}
    \caption{Qualitative segmentation results (Set 2).}
    \label{fig:supp2}
\end{figure}

\end{document}

%% file: math_commands.tex
\usepackage{amsmath,amsfonts,bm}

\def\eqref#1{equation~\ref{#1}}

\def\1{\bm{1}}

\DeclareMathAlphabet{\mathsfit}{\encodingdefault}{\sfdefault}{m}{sl}
\SetMathAlphabet{\mathsfit}{bold}{\encodingdefault}{\sfdefault}{bx}{n}

